\documentclass{tlp}
\usepackage{aopmath}

\newtheorem{definition}{Definition} 

\usepackage{graphicx}
\usepackage{enumerate}

\def\naf{{\: \mathit{not} \: }}

\begin{document}

\author{}

\title[Typed Answer Set Programming and Inverse Lambda Algorithms]{Typed Answer Set Programming Lambda Calculus Theories and
Correctness of  Inverse Lambda Algorithms with respect to them}
\author[Chitta Baral et al.]{ Chitta Baral, Juraj Dzifcak, Marcos A. Gonzalez and Aaron Gottesman\\
School of Computing, Informatics, and Decision Systems Engineering\\
Arizona State University, Tempe, AZ}

\maketitle

\begin{abstract}
Our broader goal is to automatically translate English sentences into formulas in appropriate knowledge representation languages as a step towards understanding and thus answering questions with respect to English text.  Our focus in this paper is on the language of Answer Set Programming (ASP). Our approach to translate sentences to ASP rules is inspired by Montague's use of lambda calculus formulas as meaning of words and phrases. With ASP as the target language the meaning of words and phrases are ASP-lambda formulas.  In an earlier work we illustrated our approach by manually developing a dictionary of words and their  ASP-lambda formulas. However such an approach is not scalable. In this paper our focus is on two algorithms that allow one to construct ASP-lambda formulas in an inverse manner. In particular the two algorithms take as input two lambda-calculus expressions G and H and compute a lambda-calculus expression F such that F with input as G, denoted by F@G, is equal to  H; and similarly G@F = H. We present correctness and complexity results about these algorithms. To do that we develop the notion of typed ASP-lambda calculus theories and their orders and use it in developing the completeness results.\\
\newline
\end{abstract}

\section{Introduction}

The broader goal of our proposed research is to translate English sentences to appropriate knowledge representation (KR) and reasoning languages. This will help in understanding text and answering questions with respect to it. Such an ability is important in developing various systems that need ``understanding of natural language''. This includes systems that can acquire knowledge from text, systems that can interact  in English with robots and other systems, intelligent training and tutoring systems, and systems that can process existing scientific literature in particular domains and formulate hypothesis. 

Our approach is inspired by Montague's work \cite{Montague:Book} where the meaning of words and phrases are expressed as $\lambda$-calculus expressions and the meaning of a sentence is built from the meaning of its words by making appropriate applications of the corresponding $\lambda$-calculus expressions.  This approach has also been used elsewhere, such as in \cite{Bos:Book,Collins:2005,Baral:2008,Me:2009,stefania2010,Baral:Festschrift}; but the question that we address here is  {\em how do we obtain the $\lambda$-calculus like meaning of words? They get complex quickly and hand crafting them is not scalable.} 


In this paper we use ASP as our target KR language and address the issue of automatically obtaining meaning of words as ASP-$\lambda$-calculus formulas instead of the underlying logic of first order logic in traditional use of $\lambda$-calculus.  Thus, the meanings of words are expressed as formulas of ASP-$\lambda$-calculus and using them sentences are translated to ASP rules. In \cite{Baral:2008} the ASP-$\lambda$-calculus formulas that represent words were handcrafted and it was remarked that the human engineering needed to generate the ASP-$\lambda$-calculus expressions need to be substituted by an automatic process.  

Our main idea in automating this process is through a {\em a notion of inverse application of ASP-$\lambda$-calculus formulas} and use them in constructing the ASP-$\lambda$-calculus expressions of words. We discuss two algorithms from \cite{marcos-thesis10} that compute a ASP-$\lambda$-calculus expression\footnote{This algorithm also works for typed first-order logic lambda calculus. We show that in \cite{Baral:Festschrift}. But its applicability to ASP is not discussed there as that requires additional machinery, which we present in this paper.}   $F$ given ASP-$\lambda$-calculus expressions $G$ and $H$. In the first algorithm, which we call the $Inverse_L$ algorithm, the $F$  is such that by applying $G$ as an input to it one obtains $H$; this is written as $F@G=H$. In the second algorithm, which we call the $Inverse_R$ algorithm,  $F$ is such that $G@F=H$. We refer to these algorithms as the Inverse $\lambda$-Algorithms.  In this paper we define ASP-$\lambda$-calculus formulas and formalize how this approach can be used to translate words into these formulas, yielding a method to automatically translate sentences into ASP rules.

We illustrate the basic idea of inverse application of ASP-$\lambda$-calculus formulas and how they can be used in constructing the ASP-$\lambda$-calculus expressions of words through the following example.

\begin{table}[htb]
\scriptsize{
\begin{center} 
\begin{tabular}{c c c}
Most & birds & fly  \\
$(S/(S\backslash NP))/NP$ & $NP$ & $S\backslash NP$ \\
\cline{1-2}
$S/(S\backslash NP)$   &  & $S\backslash NP$ \\
\cline{1-3}
 & & $S$ \\
\end{tabular}

\begin{tabular}{c c c}
Most & birds & fly\\
??? & $\lambda x$.$bird(x)$ & $\lambda x$.$fly(x)$ \\
\cline{1-2}
??? & & $\lambda x$.$fly(x)$ \\
\cline{1-3}
 &  & $fly(X) \leftarrow bird(X), not \neg fly(X)$ \\
\end{tabular}
\end{center}
}
\caption{CCG and $\lambda$-calculus derivation for ``Most birds fly''.}
\label{tab:MostFly1}
\end{table}

It is assumed in Table \ref{tab:MostFly1} that the meaning of ``Most birds fly'' and the ASP-$\lambda$-calculus formulas for ``fly'' and ``birds'' is known.  We would like to determine the appropriate semantic representation for the word ``most''.  To do so we must first compute the semantic representation of ``Most birds''.  This can be done using the meaning of the sentence ``Most birds fly'' and the word ``fly''.  However, we first must know whether the meaning of ``Most birds'' is to be used as input to the meaning of ``fly'' or vice versa.  To obtain this directionality information we make use of combinatory categorial grammars (CCG) \cite{Steedman:book}.

A CCG parse of a sentence assigns categories to the words of the sentence.  There are several basic categories, with $S$ representing a sentence and $NP$ representing a noun phrase.  More complex categories are formed from these basic categories by using ``$\backslash$'' and ``/'' which specify directionality.  For example, a non-transitive verb, like ``fly'' above, would have category $S\backslash NP$ meaning that if a noun phrase, $NP$, precedes the verb then a sentence $S$ is formed. Similarly, a category for a simple adjective, would be $NP/NP$, meaning that if a noun phrase,  $NP$, comes after the adjective then a  $NP$ would result.  

Note that the category of ``most'' given here is not that of a simple adjective, $NP/NP$. If ``most'' had this category then the result of applying ``birds'' to ``most'' would result in category $NP$ which would then be applied to the right of ``fly''.  However, it is not possible to form the meaning of the sentence by substituting into the given meaning of ``fly''.  Therefore, an alternative CCG parse of the sentence must be used that swaps the application of ``fly'' to be on the right side.  This is done by raising the category of ``Most birds'' to $S/(S\backslash NP)$, which in turns means that the category of ``most'' must be $(S/(S\backslash NP))\backslash NP$.

The top part of Table \ref{tab:MostFly1} gives a CCG parse of the sentence ``Most birds fly''. The meaning of the phrase ``Most birds'', which has a category $S/(S\backslash NP)$, must have the meaning of ``fly'' applied from the right since it has the category $S\backslash NP$. Therefore, to get the meaning of the sentence, $H$, we let $G$ be the meaning of ``fly''.  Then we have to find an $F$ such that $F@G=H$.  From inspection $F=\lambda x$.$(x@X \leftarrow bird(X), not \; \neg x@X)$ will satisfy this equation.

Now, having the expressions for ``Most birds'' and ``birds'', we can calculate the meaning of the desired word ``most''.   Since ``most'' has category $(S/(S\backslash NP))/NP$, we have to apply the meaning of ``birds'' to the right of it to obtain the meaning of ``Most birds''. From inspection taking the meaning of ``most'' to be $\lambda v$.$\lambda x$.$(x@X \leftarrow v@X, \; not \; \neg x@X)$ produces the desired result.

As this example demonstrates, given the meaning of most words in a sentence and a CCG parse for the sentence, we can find a new semantic representation for words and phrases whose meanings are unknown.  The question is how exactly do we determine the new representation? In this paper we discuss the Inverse-$\lambda$ Algorithms \cite{marcos-thesis10} to solve this task, which is known as the \emph{Inverse-$\lambda$ problem}. \footnote{The Inverse-$\lambda$ problem can be shown to be a special cases of the ``higher-order matching'' \cite{Dowek:Match} and ``Interpolation problem'' \cite{Stirling:Match} . Specifically, the $Inverse_L$ problem corresponds to an Interpolation problem and $Inverse_R$  problem corresponds to the Higher-order matching problem. The higher order matching problem is known to be undecidable  in the general case \cite{Loader}. Higher order matching can be further considered as a special case of higher order unification which has been explored in \cite{huet73,huet75} and recently used in \cite{kwiat10}. None of these works consider ASP.}

To help in showing the correctness and applicability of  our $Inverse_L$  and $Inverse_R$ algorithms {\em we extrapolate the notion of typed $\lambda$ first order theories  to define the notion of a typed ASP-$\lambda$ theory. We then define the notion of orders of such theories. }  Using these notions we present the soundness, completeness and complexity results of the Inverse-$\lambda$ Algorithms. For example,  the completeness result is with respect to typed ASP-$\lambda$-calculus formulas up to the second order\footnote{Blackburn and Bos say in page 101 of their book \cite{Bos:Book}: ``Now, arguably natural language semantics never requires types much above order three or so--nonetheless the ability to take a logical perspective on higher-order types really is useful.'' Note that their definition of order three corresponds to our definition of order 2.}. We then illustrate the use of the Inverse-$\lambda$ Algorithms with respect to typed ASP-$\lambda$-calculus formulas. 

As mentioned earlier, these algorithms are key to developing systems that can translate English sentences to KR languages. However, such systems need to address additional issues such as dealing with possible multiple meaning of words, and developing appropriate ontologies that maximize the accuracy of the translation. These aspects are separately discussed  in \cite{Me:2009}.  A simpler version of the Inverse-$\lambda$ Algorithms discussed in this paper is used in  developing a system that learns to translate combinatorial puzzles to ASP rules and solve those puzzles \cite{baral2012-kr}

In summary the main contributions of this paper are:

\begin{itemize}

\item We formulate the notion of typed ASP-$\lambda$-calculus theories and define  the notion of orders of such theories.

\item We illustrate the use of Inverse-$\lambda$ Algorithms with respect to typed ASP-$\lambda$-calculus formulas.

\item We present soundness, completeness and complexity results for these algorithms with respect to typed ASP-$\lambda$-calculus theories.

\end{itemize}


The rest of this paper is organized as follows. In the next section, we present some background material  and pointers on typed lambda calculus and ASP. In Section 3 we introduce typed Answer Set Programming lambda calculus. In Section 4 we present the Inverse $\lambda$-Algorithms.  We then illustrate our algorithms with respect to several examples and give a use of our algorithms in sections 5 and 6, respectively. In Section 7 we present the soundness, completeness and complexity results. Finally, we conclude and briefly mention the companion natural language semantics work that uses our algorithms.

\section{Background}

\subsection{Typed Lambda Calculus}

Since Montague's groundbreaking work \cite{Montague:Book}, $\lambda$-calculus has been accepted and used as a tool by many in natural language semantics. Montague was the first to introduce the use of $\lambda$-calculus to represent the meaning of words and $\lambda$-application as a mechanism to construct the meaning of phrases and sentences. However, to properly understand the notion of ``meaning'' (or semantics), it is useful to consider models of $\lambda$-calculus expressions. When referring to a model, one is looking for a semantic tool that can give it two elements: the entities that are part of the domain, and for every element in the signature, the semantic value associated with it. By creating this model with the corresponding denotations for types, expressions of the system will have a defined type and a semantic value associated with it.

Both untyped and typed $\lambda$-calculus can be characterized using models, but typed $\lambda$-calculus has had the most impact on natural language semantics, which also became familiar to linguistics after the mentioned works by Montague. In this paper we will follow the \textit{Simply Typed Lambda Calculus} of Church \cite{Church:1940} to have ASP as the core logic. This is the most commonly used approach in linguistics where only one type constructor is used to build types, ``$\rightarrow$'', and each term has a single type associated with it \cite{Barendregt:Book}.

Because of space constraints, we do not present the typed lambda calculus definitions that we use to define typed ASP-$\lambda$ calculus. The books \cite{Hindley:Book} and \cite{HindleyType:Book} are good reference points for typed lambda calculus.

\subsection{Answer Set Programming}

Answer Set Programming is the language of logic programming with answer set semantics \cite{Gelfond-Lifschitz}. This language is one of the most suitable declarative languages \cite{Baral:Book} for knowledge representation, reasoning and declarative problem solving; all important aspects that are needed to develop natural language understanding systems. It has a large body of support structure, including efficient implementations and theoretical building block studies. It allows the representation, in an intuitive way, of various kinds of knowledge that cannot be adequately expressed in first-order logic. These include, for instance, default statements ($most$ $birds$ $fly$) and normative statements ($normally$ $birds$ $fly$).  We now present some basic definitions related to Answer Set Programming syntax and semantics \cite{Baral:Book}.

\begin{definition}[ASP rule]
An \textit{ASP rule} is of the form:
\begin{center}
$L_0 \; or  \; $\ldots$ \; or \; L_k \; \leftarrow \; L_{k+1} \; , \; $\ldots$ \; , L_m \; , \; not \; L_{m+1} \; , \; $\ldots$ \; not \; L_n.$
\end{center}
where \textit{$L_i$} are literals and k $\geq$ 0, m $\geq$ k and n $\geq$ m.
\end{definition}

The literals to the left of the ``$\leftarrow$'' belong to the \textit{Head} of the rule, and the literals to the right of the ``$\leftarrow$'' belong to the \textit{Body} of the rule. An \textit{ASP program} is a set of \textit{ASP rules}.

\begin{definition}[Satisfiability]
An \textit{ASP rule} of the form:
\begin{center}
$L_0  \; or \; $...$. \; or \; L_k \leftarrow \; L_{k+1} \; , \; $...$ \; , L_m \; , \; not \; L_{m+1} \; , \; $...$ \; not \; L_n.$
\end{center}
of an ASP program $\Pi$ is said to be \textit{satisfied} by a set of ground literals \textit{I} of $\Pi$ if: \\
$\bullet$ 
\{$L_{k+1} \; , $...$ , L_m$\} $\subseteq$ \textit{I} and \{$L_{m+1} \; , $...$ , \; L_n$\} $\cap$ \textit{I} = $\emptyset$ implies that \{$L_0,$...$,L_k$\} $\cap$ \textit{I} $\neq$ $\emptyset$.
\end{definition}

\begin{definition}
An \textit{Answer Set} of an ASP Program $\Pi$ without the ``not'' operator, is a consistent set of ground literals $S$ such that $S$ satisfies all rules of $\Pi$ and no subset of $S$ satisfies all rules of $\Pi$ .
\end{definition}

\begin{definition}[Answer Set]
A consistent set $S$ of ground literals is an \textit{Answer Set} of an ASP Program $\Pi$ if $S$ is an answer set of the reduct $\Pi^S$, where $\Pi^S$ is obtained from $\Pi$ by
\begin{enumerate}
\item[(i)] Deleting all rules from $\Pi$ that contain some $\naf l$ in their body for some $l \in S$.
\item[(ii)] Removing all occurrences of $\naf l$ from the remaining rules.
\end{enumerate}
\end{definition}

\section{Typed Answer Set Programming Lambda Calculus}

We start by presenting the signature for the language Typed Answer Set Programming Lambda Calculus (Typed ASP Lambda Calculus). It consists of the following:
\begin{itemize}
\item the lambda operator, also called abstractor, $\lambda$;
\item the lambda application $@$;
\item the parenthesis  symbols $(, ), [,$ and $]$;
\item for every type $a$, an infinite set of variables $v_{n,a}$ for each natural number $n$;
\item for every type $a$, a (possibly empty) set of constants $c_{a}$ of type $a$;
\item the connectives $or$ , $\; \leftarrow \;$, $\neg$ , $not$, ``,'' and ``.''; and
\item predicate and function symbols.
\end{itemize}
Variables and constants in the signature for Typed ASP Lambda Calculus will be referred to as \emph{$\lambda$-terms}.


Next, we introduce the set of types that will be used with Typed ASP Lambda Calculus, in conjunction with the definition of the semantics of types assigned to the different expressions of the language. We will follow the principles presented in \cite{MathLing:Book}, where $D_a$ represents the set of possible objects (\textit{denotations}) that describe the meanings of expressions of type a.

\begin{definition}[Types]
The \textit{set of types} $\Theta$ is defined recursively as follows:
\begin{enumerate}
\item  $e,a,l,g,d,h,t$ are types, called \emph{base types}, and
\item if $A$ and $B$ are types, then $(A \rightarrow B)$ is a type.
\end{enumerate}
\end{definition}

Intuitively, $e$ refers to terms, which is either a variable or a constant in ASP, or a function symbol with terms as input; $a$ refers to atoms of ASP, which are predicate symbols with terms as input\footnote{An atom is said to be ground if none of the terms in the atom contain a variable}; $l$ refers to literals of ASP which are atoms or atoms preceded by the connective $\neg$; $g$ refers to \textit{gen-literals} which are literals or literals preceded by the connective $not$; $d$ refers to a conjunction of gen-literals, where the conjunction is denoted by ``,''; $h$ refers to a disjunction of literals, where the disjunction is denoted by ``$or$''; and $t$ refers to the boolean truth values. More formally,  

\begin{definition}[Type Semantics]
Given an ASP Program $\Pi$, the semantics of $\Pi$ is defined using:
\begin{itemize}
\item $D_{e}$ = the set of terms and functions in the language of $\Pi$;
\item $D_{a}$ = the set of atoms in the language of $\Pi$;
\item $D_{l}$ = the set of literals;
\item $D_{g}$ = the set of gen-literals;
\item $D_{h}$ = the set of ``$or$''-connected literals that belong to heads of ASP rules; 
\item $D_{d}$ = the set of  ``$,$''-connected gen-literals that belong to bodies of ASP rules;
\item $D_{t} = \{0,1\}$, the set of satisfiability values for an ASP program;  and
\item $D_{a \rightarrow b} =$ the set of functions from $D_{a}$ to $D_{b}$.
\end{itemize}
\end{definition}

Expressions of type $t$ denote satisfiability values of ASP programs. An ASP program can be true under certain Herbrand interpretations, and false under others. $(a \rightarrow b)$ denotes functions whose input is in $D_{a}$ and output values are in $D_{b}$. For example, the type $(e \rightarrow t)$ corresponds to functions from terms to satisfiability values.

This section continues by introducing the definition for ASP typed term, followed by the definition of ASP $\lambda$-calculus formula:

\begin{definition}[ASP Typed Term]
The elements which belong to the set $\Delta_{A}$ of ASP typed terms of type $A$ are inductively defined as follows:
\begin{enumerate}
\item For each type $A$, every $\lambda$-term of type $A$ belongs to $\Delta_{A}$.
\item For any types $A$ and $B$ 
\begin{itemize}
\item if $\alpha$ $\in$ $\Delta_{A \rightarrow B}$ and $\beta$ $\in$ $\Delta_{A}$, then $\alpha$@$\beta$ $\in$ $\Delta_{B}$
\item if $u$ is a variable of type $A$ and $\alpha$ $\in$ $\Delta_{B}$ has free occurrences of the variable $u$, then $\lambda$u.$\alpha$ $\in$ $\Delta_{A \rightarrow B}$ and the free occurrences of $u$ are now bound to the abstractor $\lambda$u.\footnote{Refer to the definition of occurrence presented at the end of this section.}
\end{itemize}
\item If $f$ is a function symbol with arity $n$, and $t_1,t_2,\ldots,t_n \in \Delta_{e}$, then \\$f(t_1,t_2,\ldots,t_n) \in \Delta_{e}$.
\item If $p$ is a predicate symbol with arity $n$, and $t_1,t_2,\ldots,t_n \in \Delta_{e}$, then \\$p(t_1,t_2,\ldots,t_n) \in \Delta_{a}$.
\item If $\alpha \in \Delta_{a}$ and $\alpha$ is not a $\lambda$-term\footnote{This is to guarantee that each $\lambda$-term only corresponds to its unique type.}, then $\alpha \in \Delta_{l}$ and  ($\neg \alpha) \in \Delta_{l}$.
\item If $\alpha \in \Delta_{l}$ and $\alpha$ is not a $\lambda$-term, then $\alpha \in \Delta_{g}$ and  ($not$ $\alpha) \in \Delta_{g}$.
\item If $\alpha \in \Delta_{l}$ and $\alpha$ is not a $\lambda$-term, then $\alpha \in \Delta_{h}$.
\item If $\alpha,\beta  \in \Delta_{h}$, then $\alpha$ $or$ $\beta \in \Delta_{h}$.
\item If $\alpha \in \Delta_{g}$ and $\alpha$ is not a $\lambda$-term, then $\alpha \in \Delta_{d}$.
\item If $\alpha,\beta  \in \Delta_{d}$, then $\alpha, \beta \in \Delta_{d}$.
\item If $\alpha \in \Delta_{h}$ and $\beta \in \Delta_{d}$, then ($\alpha \leftarrow .) \in \Delta_{t}$, ($ \leftarrow \beta.) \in \Delta_{t}$, and ($\alpha \leftarrow \beta.) \in \Delta_{t}$.
\item If $\rho_1, \rho_2 \in \Delta_{t}$, then $(\rho_1 \; \rho_2) \in \Delta_{t}$.
\end{enumerate}
\end{definition}

\begin{definition}[Typed ASP $\lambda$-Calculus Formula]\label{defn-8}
A typed ASP $\lambda$-calculus formula is an ASP typed term where every variable is bound to an abstractor and every abstractor binds to a variable.
\end{definition}

A typed ASP $\lambda$-calculus formula is in $\beta$-normal form if it does not contain any $\beta$-redex occurrences. An example of a $\beta$-redex is a typed term of the form ($\lambda$v.v)@$John$. The typed term $John$ or $\lambda$v.v, do not have any $\beta$-redex occurrences \cite{HindleyType:Book}.



The binding of every variable to an abstractor and every abstractor to a variable in definition~\ref{defn-8}  correspond to closed and $\lambda$I-terms in the classic theory of lambda calculus, respectively. These conditions ensure that one obtains Answer Set Programming programs when the typed ASP $\lambda$-calculus formulas are in $\beta$-normal form. By the way in which the ASP typed terms have been defined and the two properties that we are enforcing, when there are no lambda abstractors left in an ASP typed term we obtain an expression that belongs to the Answer Set Programming language presented above. Some examples of typed ASP $\lambda$-calculus formulas are the following:

\begin{itemize}
\item $\lambda w$.$ \lambda v$.$(w \; \leftarrow \; v@X$.$)$ with type $(h \rightarrow ((e \rightarrow d) \rightarrow t))$ where $w$ has type $h$, $v$ has type $(e \rightarrow d)$, and $X$ has type $e$.
\item $\lambda x$.$ \lambda y$.$(\leftarrow \; h(x), \, not \, \neg\, y$.$)$ with type $(e \rightarrow (a \rightarrow t))$ where $x$ has type $e$ and $y$ has type $a$.
\item $\lambda v$.$ (v \, or \, \neg \, v \leftarrow.)$ with type $(a \rightarrow t)$ where $v$ has type $a$.
\item $\lambda w$.$(\lambda u$.$( w@ \lambda v$.$(position(v,u))))$ with type $(((e \rightarrow l) \rightarrow t) \rightarrow (e \rightarrow t))$ where $w$ has type $((e \rightarrow l) \rightarrow t)$, $u$ and $v$ have type $e$.
\end{itemize}





Let the fourth formula of the examples be $J$. In $J$, $w$ has type $((e \rightarrow l) \rightarrow t)$ because when an ASP typed formula is applied $J$, it will be placed in the variable $w$ and will receive as argument the expression $\lambda v$.$(position(v,u))$. This expression has type $(e \rightarrow l)$ and therefore the input of the formula applied to $J$ needs to have $(e \rightarrow l)$ as input and $t$ as output to lead to an ASP formula. Thus, $w$ has type $((e \rightarrow l) \rightarrow t)$. $u$ has type $e$ meaning that one expects a term to be placed inside the literal $position$.

The following are \textbf{not} typed ASP $\lambda$-calculus formulas:

\begin{enumerate}
\item $\lambda y$.$\lambda x$.$(y \; or \; not \, x@X)$, where $x$ has type $(e \rightarrow l)$ and $y$ have type $l$.
\item $\lambda v$.$ \lambda w$.$(\neg \, w \; \leftarrow \; \neg \, not \, v@X)$, where $w$ has type $a$, $v$ has type $(e\rightarrow l)$, and $X$ has type $e$.
\end{enumerate}

The first expression is not a $\lambda$-calculus formula since $x$ is of type $(e \rightarrow l)$ which means $x@X$ has type $l$ from $r_2$ above.  Then $not \, x@X$ must have type $g$ from $r_5$.  However, there is no rule that allows us to combine $y$, an element of type $l$, with $not \, x@X$, an element of  type $g$, with the connective $or$.

The second expression violates the rules of a typed ASP $\lambda$-calculus formulas since there is no rule saying that the connective $\neg$ can be applied to terms of type $g$, which in this example is the type of $not \, v@X$. 




This section concludes with two more definitions.

\begin{definition}[Occurrence]
The relation \emph{P occurs in Q} is defined by induction on $Q$ as follows:
\begin{itemize}
\item an ASP typed term $P$ occurs in $P$.
\item if $P$ occurs in $M$ or in $N$, then $P$ occurs in $M@N$.
\item if $P$ occurs in $M$, then $P$ occurs in $\lambda x.M$. 
\item if $P$ occurs in $\phi$ or $P$ occurs in $\psi$, then $P$ occurs in $\phi$ $or$ $\psi$, $\phi \leftarrow \psi$ and $\phi$ $,$ $\psi$.
\item if $P$ occurs in $\phi$, then $P$ occurs in $\neg \phi$ and $not$ $\phi$.
\item if $P$ occurs in any term $t_i$, then $P$ occurs in $F(t_{1},$...$,t_{n})$. Where $F$ is a function symbol.
\item if $P$ occurs in any term $t_i$, then $P$ occurs in $R(t_{1},$...$,t_{n})$. Where $R$ is a predicate symbol of an atom.
\end{itemize}
\end{definition}

\begin{definition}[sub-term]
A sub-term of a typed ASP $\lambda$-calculus formula $F$ is any term $P$ that occurs in $F$.
\end{definition}

\subsection{Type Order}

We have introduced the types that will be assigned to typed ASP lambda calculus terms and formulas. Next, we present the notion of order, which is associated with types and establishes a hierarchical structure that separates typed $\lambda$-calculus formulas to several classes. Order will be an important concept when we state the completeness proof for the Inverse $\lambda$-Algorithms since we will show that they are complete for typed $\lambda$-calculus formulas up to order two. Each typed term has a type, and each type will be assigned an order.

\begin{definition}[Type Order]
The order of a type is defined as:
\begin{enumerate}
\item Base types have order 0.
\item For function types, $order(a \rightarrow  b$) = $max(order(a)+1, order(b))$.
\end{enumerate}
\end{definition}

The definition from \cite{Stirling:Match} gives order one to base types. In our case, we consider order zero for base types since this is the common approach in linguistics. 
Next, some examples of typed ASP lambda calculus formulas with different orders are presented:

\begin{itemize}
    \item Order zero: $bird(tweety).$ - type $t$.
	 \item Order one: $\lambda v$.$\lambda u$.$(v \leftarrow u.)$ - type $(h \rightarrow (d \rightarrow t))$.
	 \item Order two: $\lambda v$.$\lambda u$.$(v@X \leftarrow u@X.)$ - type $((e \rightarrow l) \rightarrow ((e \rightarrow g) \rightarrow t))$.
	 \item Order three: $\lambda w$.$(w@(\lambda z$.$h(z)).)$ - type $(((e \rightarrow l) \rightarrow t) \rightarrow t)$.
\end{itemize}

With these simple examples, one can see the intuition behind the order of typed ASP lambda calculus formulas. Formulas of order zero correspond to expressions with base types. Formulas of order one correspond to expressions which start with a series of lambda abstractors followed by an ASP program with variables bound to the initial lambda abstractors.

Formulas of order two extend the expressions allowed in order one by including applications. Formulas of order zero can be applied to variables inside the formula. Formulas of order three extend those present in order two by allowing lambda abstractors inside the expression after the initial lambda abstractors. In this case, formulas of order one can be applied to variables, this is why now, we can find lambda abstractors at the beginning and in the middle of the formulas. These claims can be easily proved by contradiction using the given definitions. 

\section{The Inverse Lambda Operators}

This section presents the formal definition of the two components of the Inverse $\lambda$-Algorithms, $Inverse_L$ and $Inverse_R$, from \cite{marcos-thesis10}. The objective of $Inverse_L$ and $Inverse_R$ is that, given typed $\lambda$-calculus formulas $H$ and $G$, the formula $F$ is computed such that $F@G=H$ and $G@F=H$, respectively. We now define the different symbols used in the algorithm and their meaning:

\begin{itemize}
\item Let $G$, $H$ and $J$ represent typed $\lambda$-calculus formulas, $J^1$,$J^2$,...,$J^n$ represent typed terms; $v$, $w$ and $v_1$,...,$v_n$ represent variables.
\item Typed terms that are sub-terms of a typed term $J^i$ are denoted as $J^i_k$.
\end{itemize}

We also consider the following two statements:

\begin{itemize}
\item A list of $\lambda$-abstractors of the form $\lambda v_1,$...$,v_i$ can be empty if the corresponding variables $v_1,$...$,v_i$ are not present in the formula they belong to.
\item If the formulas being processed within the algorithm do not satisfy any of the $if$ conditions then the algorithm returns $null$.
\end{itemize}

\begin{definition}[Operator :]
Consider two lists (of same length) of typed ASP $\lambda$-calculus formulas $A_1,$...$,A_n$ and $B_1,$...$,B_n$, and a typed ASP $\lambda$-calculus formula $H$. The result of the operation $H(A_1,$...$,A_n:B_1,$...$,B_n)$ is defined as:
\begin{enumerate}
    \item find the first occurrence of formulas $A_1,$...$,A_n$ in $H$.
	 \item replace each $A_i$ by the corresponding $B_i$.
	 \item find the next occurrence of formulas $A_1,$...$,A_n$ in $H$ and go to 2. Otherwise, stop.
\end{enumerate}
\end{definition}

We now give the two inverse algorithms.

\begin{definition}[$Inverse_L(H,G)$]
The algorithm $Inverse_L(H,G)$, is defined as:
\newline
\noindent Given $G$ and $H$:
\begin{it}
\begin{enumerate}
\item If  $G$ is $\lambda v$.$v$
\begin{itemize}
\item  then $F$ = $\lambda v$.$(v@H)$
\end{itemize}
\item If $G$ is a sub-term of $H$
\begin{itemize}
\item then $F$ = $\lambda v$.$H(G:v)$
\end{itemize}
\item If $G$ is not $\lambda v$.$v$, $(J^1(J^1_1,$...$,J^1_m), J^2(J^2_1,$ ... $,J^2_m),\;$ ... $\;,J^n(J^n_1,$...$,J^n_m))$ are sub-terms of $H$, and $\forall J^i \in H$, $G$ is $\lambda v_1,$...$,v_s.J^i(J^i_1,$...$,J^i_m:v_{k_1},$...$,v_{k_m})$\footnote{When the formula $G$ is being generated, the indexes of the abstractors $\lambda v_1,$...$,v_s$ must be assigned to bind the variables from $v_{k_1},$...$,v_{k_m}$ in such a way that $G$ is a valid formula.} with 1 $\leq$ $s$ $\leq$ $m$ and $\forall$$p$, 1 $\leq$ $k_p$ $\leq$ $s$.
\begin{itemize}
\item then $F$ = $\lambda w$.$H((J^1:(w@J^1_{k_1}@$...$@J^1_{k_m}),$...$,J^n:(w@J^n_{k_1}@$ ... $@J^n_{k_m})))$ where each $J_{k_p}$ maps to a different $v_{k_p}$ in $G$.
\end{itemize}
\item If $H$ is $\lambda v_1,$...$,v_i.J$ and $J^1(J^1_{i+1},$ ... $,J^1_{s})$ is a sub-term of $J$,\newline $G$ is $\lambda w$.$J(J^1(J^1_{i+1},$...$,J^1_{s}):w@J^1_{k_1}@$...$@J^1_{k_s})$ with $\forall$$p$, $i+1$ $\leq$ $k_p$ $\leq$ $s$.
\begin{itemize}
\item then $F$ = $\lambda w$.$ \lambda v_1,$...$,v_i.(w@\lambda v_{i+1},$...$,v_s.(J^1(J^1_{i+1},$...$,J^1_{s}:v_{k_1},$...$,v_{k_s})))$
\end{itemize}

\end{enumerate}
\end{it}
\end{definition}

\begin{definition}[$Inverse_R(H,G)$]
The algorithm $Inverse_R(H,G)$, is defined as:
\newline
\noindent Given $G$ and $H$:
\begin{it}
\begin{enumerate}
\item If $G$ is $\lambda v$.$v@J$
\begin{itemize}
\item then $F = Inverse_L(H,J)$
\end{itemize}
\item If $J$ is a sub-term of $H$ and $G$ is $\lambda v$.$H(J:v)$
\begin{itemize}
\item  then $F$ = $J$
\end{itemize}
\item If $G$ is not $\lambda v$.$v@J$, $(J^1(J^1_1,$...$,J^1_m), J^2(J^2_1,$...$,J^2_m),\;$ ... $\;,J^n(J^n_1,$...$,J^n_m))$ are sub-terms of $H$ and $G$ is $\lambda w$.$H((J^1(J^1_1,$...$,J^1_m):w@J^1_{k_1}@$...$@J^1_{k_m}),$...$,(J^n(J^n_1,$ ... $,J^n_m):w@J^n_{k_1}@...@J^n_{k_m}))$ with 1 $\leq$ $s$ $\leq$ $m$ and $\forall$$p$, 1 $\leq$ $k_p$ $\leq$ $m$.
\begin{itemize}
\item then $F$ = $\lambda v_1,$...$,v_s$.$J^1(J^1_1,$...$,J^1_m:v_{k_1},$...$,v_{k_m})$.
\end{itemize}
\item If $H$ is $\lambda v_1,$...$,v_i$.$J$ and $J^1(J^1_{i+1},$...$,J^1_{s})$ is a sub-term of $J$, \newline $G$ is $\lambda w$.$\lambda v_1,$...$,v_i$.$(w@\lambda v_{i+1},$...$,v_s.(J^1(J^1_{i+1},$...$,J^1_{s}:v_{k_1},$...$,v_{k_s})))$ with $\forall$$p$, $i+1$ $\leq$ $k_p$ $\leq$ $s$.
\begin{itemize}
\item then $F$ = $\lambda w.J(J^1(J^1_{i+1},$...$,J^1_{s}):w@J^1_{k_1}@$...$@J^1_{k_s})$
\end{itemize}
\end{enumerate}
\end{it}
\end{definition}

Please note that the final cases for both operators involve formulas of third order. 

\section{Inverse Lambda Algorithm Examples with Typed ASP Lambda Calculus}

This section presents several examples demonstrating how the Inverse-$\lambda$ Algorithms can be applied to find $F$ in various settings given ASP-$\lambda$-calculus formulas $G$ and $H$ .  A use case example follows in the next section.

\subsection{Example 1}

Let $H$ and $G$ be typed ASP $\lambda$-calculus formulas where $H$ = $bird(tweety)$. and $G$ = $\lambda$$x$.$x$. $F$ needs to be calculated such that $H$ = $F$ @ $G$. Here case 1 of $Inverse_L$ will be applicable. Then $F$ = $\lambda$$v$.$(v$ @ $H)$ and in this case $F$ = $\lambda$$v$.$(v$ @ $bird(tweety))$. Then, $F$ @ $G$  = $\lambda$$v$.$(v$ @ $bird(tweety))$ @ $\lambda$$x$.$x$ = ($\lambda$$x$.$x$ @ $bird(tweety))$ = $bird(tweety)$ = $H$.





\subsection{Example 2}

Let $H$ and $G$ be typed ASP $\lambda$-calculus formulas where $H$ = $\lambda u$.$ (fly(X)$ $\leftarrow$ $u,$ $not$ $\neg fly(X)$.$)$ and $G$ = $fly(X)$. $F$ needs to be calculated such that $H$ = $F$ @ $G$. Here case 2 of $Inverse_L$ is applicable. Thus, we get $F$ = $\lambda v$.$H(G$ : $v)$  = $\lambda v$.$H(fly(X)$ : $v)$ = $\lambda v$.$\lambda u$.$ (v$ $\leftarrow$ $u,$ $not$ $\neg v$.$)$






\subsection{Example 3}

Let $H$ and $G$ be typed ASP $\lambda$-calculus formulas with $H$ = $\lambda u$.$ (bird(tweety),$ $animal(tweety),$ $penguin(rocky),$ $animal(rocky),$ $eats(tweety,u))$ and $G$ = $\lambda v$.$ \lambda w$.$ (v,$ $animal(w))$.  $F$ now needs to be calculated such that $H$ = $F$ @ $G$. Therefore, case 3 of $Inverse_L$ will be applicable.

$G$ is not $\lambda v$.$v$ so the first condition is satisfied. From $H$, one has the following formulas that are subterms: $J^1$ = $bird(tweety),$ $animal(tweety)$ with sub-subterms $J^1_1$ = $bird(tweety)$ and $J^1_2$ = $tweety$ (from $animal(tweety)$); $J^2$ = $penguin(rocky),$ $animal(rocky)$ with sub-subterms $J^2_1$ = $penguin(rocky)$ and $J^2_2$ = $rocky$ (from $animal(rocky)$).  Therefore the second condition of case 2 is satisfied.

The third condition is satisfied since, $\forall J^i$ $\in$ $H$: $G$ = $\lambda v_1$.$\lambda v_2$.$J^i(J^i_1,$ $J^i_2$ : $ v_1$, $v_2)$ for $i$ = 1,2. For example, for $J^1$, $G$ =  $\lambda v_1$.$\lambda v_2$.$J^1(bird(tweety)$, $tweety$ : $v_1$, $v_2)$ = $\lambda v$.$\lambda w$.$(v,$ $animal(w))$.

Therefore one can now calculate that $F$ = $\lambda w$.$H((J^1$ : $w$ @ $J^1_1$ @ $J^1_2),$ $(J^2$ : $w$ @ $J^2_1$ @ $J^2_2))$ = $\lambda x$.$H((J^1$ : $x$ @ $bird(tweety)$ @ $tweety),$ $(J^2$ : $x$ @ $penguin(rocky)$ @ $rocky))$ = $\lambda x$.$\lambda u$.$(x$ @ $bird(tweety)$ @ $tweety$, $x$ @ $penguin(rocky)$ @ $rocky,$ $eats(tweety,u))$.

\subsection{Example 4}

Let $H$ and $G$ be typed ASP $\lambda$-calculus formulas with $H$ = $love(Mia$, $Jon)$ $\leftarrow$ $love(Jon$, $Mia)$. and $G$ = $\lambda w$.$w$ @ $Mia$ @ $Jon$ $\leftarrow$ $w$ @ $Jon$ @ $Mia$.  $F$ needs to be calculated such that $H$ = $G$ @ $F$. Case 3 of $Inverse_R$ will be applied.

$G$ is clearly not $\lambda v$.$v$ @ $J$.  $H$ has the following subterms: $J^1(J^1_{1},$ $\ldots$, $J^1_{m})$ = $love(Mia$, $Jon)$ with sub-subterms $J^1_1$ = $Mia$ and $J^1_2$ = $Jon$; $J^2(J^2_{1}$, $\ldots$, $J^2_{m})$ = $love(Jon$, $Mia)$  with $J^2_1$ = $Jon$ and $J^2_2$ = $Mia$. Then, $G$ =$\lambda w$.$H((J^1(J^1_1$, $J^1_2)$ : $w$ @ $J^1_1$ @ $ J^1_2)$,  $(J^2(J^2_1$, $J^2_2)$ : $w$ @ $J^2_1$ @ $J^2_2))$ = $\lambda w$.$(love(Mia$, $Jon)$ : $w$ @ $Mia$ @ $Jon)$ $\leftarrow$ $(love(Jon$, $Mia)$ : $w$ @ $Jon$ @ $Mia)$. = $\lambda w$.$w$ @ $Mia$ @ $Jon$ $\leftarrow$ $w$ @ $Jon$ @ $Mia$. Therefore, $G$ satisfies the second condition of case 3. 

Thus, we calculate $F$ =$\lambda v_1$.$\lambda v_2$.$J^1(J^1_{1}$, $J^1_2$ : $v_1$, $v_2)$ = $\lambda v_1$.$\lambda v_2$.$(love(Mia$, $Jon$ : $v_1$, $v_2))$ = $\lambda v_1$.$\lambda v_2$.$love(v_1$, $v_2)$.






\subsection{Example 5}

Let $H$ and $G$ be typed ASP $\lambda$-calculus formulas where $H$ = $\lambda v$.$(stay\_at(room5)$ $\leftarrow$ $not$ $goto\_from(v$, $room5)$.$)$ and $G$ = $\lambda w$.$ \lambda v$.$(w$ @ $\lambda u$.$goto\_from(v$, $u))$. $F$ needs to be calculated such that $H$ = $G$ @ $F$. Therefore, case 4 of $Inverse_R$ will be applied.

$H$ = $\lambda v$.$J$ with $J$ = $stay\_at(room5)$ $\leftarrow$ $not$ $goto\_from(v,$ $room5)$. $f(\sigma_{i+1}$, $\ldots$, $\sigma_s$) = $goto\_from(v,room5)$  with $s$ = 2 and $\sigma_2$ = $room5$. Relabeling the variables of $G$ to better match the conditions of the case by substituting $v_1$ for $v$ and $v_2$ for $u$, we see $G$ = $\lambda w$.$\lambda v_1$.$(w$ @ $\lambda v_2$.$ (f(\sigma_{2}$ : $v_{2})))$. Therefore, $G$ satisfies the second condition of case 4.

Thus, we calculate $F$ = $\lambda w$.$J(f(\sigma_{2})$ : $w$ @ $\sigma_2)$ = $\lambda w$.$(stay\_at(room5)$ $\leftarrow$ $not$ $w$ @ $room5$.$)$






\section{Use Case Examples}

In this section we present a use case of our inverse lambda algorithms to show how meaning of words are computed when one knows meaning of the sentences and meaning of some of the words. We consider the following sentences from  \cite{Baral:2008}. 

\begin{itemize}
	\item Most birds fly.
	\item Penguins are birds.
	\item Penguins do not fly.
\end{itemize}

We will consider an initial lexicon that has the semantics for simple nouns and verbs. Combinatory Categorial Grammar (CCG) \cite{Clark:CCG} is used to construct the meaning of a sentence from the meaning of its constituent words and phrases. After parsing the first two sentences using CCG and adding the semantics from the initial lexicon, we obtain the output of a simplified CCG parsing with two categories ``S''(sentence) and ``NP''(noun phrase), as shown in Table \ref{tab:MostFly}. \medskip

\begin{table}[htb]
\tiny{
\begin{center} 

\begin{tabular}{c c c}
Most & birds & fly  \\
$(S/(S\backslash NP))/NP$ & $NP$ & $S\backslash NP$ \\
\cline{1-2}
$S/(S\backslash NP)$   &  & \\
\cline{1-3}
 & & $S$ \\
\end{tabular}

\begin{tabular}{c c c}
Penguins & are & birds\\
$NP$ & $(S\backslash NP) / NP$ & $NP$\\
\cline{2-3}
$NP$ & $S\backslash NP$ & \\
\cline{1-3}
 & $S$ & \\
\end{tabular}

\begin{tabular}{c c c}
Most & birds & fly\\
??? & $\lambda x$.$bird(x)$ & $\lambda x$.$fly(x)$ \\
\cline{1-2}
??? & & $\lambda x$.$fly(x)$ \\
\cline{1-3}
 &  & $fly(X) \leftarrow bird(X), not \neg fly(X)$ \\
\end{tabular}

\begin{tabular}{c c c}
Penguins & are & birds\\
$\lambda x$.$penguin(x)$ & ??? & $\lambda x$.$bird(x)$\\
\cline{2-3}
$\lambda x$.$penguin(x)$ & ??? &\\
\cline{1-3}
 & $penguin(X) \leftarrow bird(X)$ & \\
\end{tabular}
\end{center}
}
\caption{CCG and $\lambda$-calculus derivation for ``Most birds fly'' and ``Penguins are birds''.}
\label{tab:MostFly}
\end{table}

In Table \ref{tab:MostFly}, one can see that the semantic representations for the words ``most'' and ``are'' are missing. We already discussed how we can obtain the semantic representation of ``most'' before. Now we will illustrate how we can compute it using the presented Inverse $\lambda$-Algorithms. Starting with the first sentence, one can take the meaning of the sentence and the meaning of the word ``fly'' to calculate the representation of ``Most birds''.

``Most birds'' has category $S/(S\backslash NP)$ and the category of ``fly'' is being applied from its right. Therefore, if we take $H$ as the meaning of the sentence and $G$ as the meaning of ``fly'', we can use $Inverse_L(H,G)$ to obtain the expression for ``Most birds''. In this case, option three of the algorithm is satisfied and $F$ = $\lambda x$.$(x@X \leftarrow bird(X), not \; \neg x@X)$.

Now, we have the expression for ``Most birds'' and ``birds''. Since the word ``birds'' is being applied to the right of ``Most birds'', we need to again call $Inverse_L(H,G)$ to obtain the representation for ``Most''. Option three of the algorithm is again satisfied and one obtains $F$ as $\lambda v$.$\lambda x$.$(x@X \leftarrow v@X, \; not \; \neg x@X)$. This is the Typed ASP $\lambda$-calculus representation for the word ``most''.

The process to obtain the word ``are'' from the second sentence is very similar. First one calls $Inverse_L(H,G)$ with $H$ being the meaning of the sentence and $G$ being ``Penguins'' to obtain the meaning of ``are birds''. Option three of the algorithm is satisfied and $F$ = $\lambda x$.$(x@X \leftarrow bird(X))$. Next, one calls $Inverse_L(H,G)$ with ``are birds'' and ``birds'' to obtain the desired meaning of ``are''. Option three of the algorithm is satisfied again and $F$ = $\lambda v$.$\lambda x$.$(x@X \leftarrow v@X$). This expression corresponds to the Typed ASP $\lambda$-calculus formula for the word ``are''. Next, the last sentence is presented in Table \ref{tab:PengSwim}.

\begin{table}[htb]
\tiny{
\begin{center} 


\begin{tabular}{c c c}
Penguins & do not & fly\\
$NP$ & $(S/(S\backslash NP))\backslash NP$ & $S\backslash NP$\\
\cline{1-2}
$(S/(S\backslash NP))$ & $S\backslash NP$\\
\cline{1-3}
 & $S$ & \\
\end{tabular}


\begin{tabular}{c c c}
Penguins & do not & fly\\
$\lambda x$.$penguin(x)$ & ??? & $\lambda x$.$fly(x)$\\
\cline{1-2}
& ??? & $\lambda x$.$fly(x)$\\
\cline{1-3}
 & $\neg fly(X) \leftarrow penguin(X)$ & \\
\end{tabular}
\end{center}
}
\caption{CCG and $\lambda$-calculus derivations for ``Penguins do swim'' and ``Penguins do not fly''.}
\label{tab:PengSwim}
\end{table}

In this case, the semantics of the phrase ``do not'' is missing. This phrase was not part of the initial lexicon. 
For this sentence, we call $Inverse_L$ first, obtaining $\lambda x$.$(\neg x@X \leftarrow penguin(X))$ as the meaning of ``Penguins do not'', and $Inverse_L$ afterwards to obtain the representation for ``do not'', which is $\lambda u$.$\lambda x$.$(\neg x@X \leftarrow u@X$).

\section{Correctness and Complexity of the Inverse Algorithms}

\begin{theorem}[Soundness of $Inverse_L$]
Given two typed $\lambda$-calculus formulas $H$ and $G$ in $\beta$-normal form, if $Inverse_L(H,G)$ returns a non-null value $F$, then
\noindent $H$ = $F$ $@$ $G$.
\end{theorem}

\begin{theorem}[Soundness $Inverse_R$]
Given two typed $\lambda$-calculus formulas $H$ and $G$ in $\beta$-normal form, if $Inverse_R(H,G)$ returns a non-null value $F$, then
\noindent $H$ = $G$ $@$ $F$.
\end{theorem}

\begin{theorem}[Completeness of $Inverse_L$]
For any two typed $\lambda$-calculus formulas $H$ and $G$ in $\beta$-normal form, where H is of order two or less, and G is of order one or less, if there exists a set of typed $\lambda$-calculus formulas $\Theta_F$ of order two or less in $\beta$-normal form, such that $\forall F_i$ $\in$ $\Theta_F$, $H = F_i @ G$, then $Inverse_{L}(H,G)$ will give an $F$ where $F$ $\in$ $\Theta_F$.
\end{theorem}

\begin{theorem}[Completeness of $Inverse_R$]
For any two typed $\lambda$-calculus formulas $H$ and $G$ of order two or less in $\beta$-normal form, if there exists a set of typed $\lambda$-calculus formulas $\Theta_F$ of order one or less in $\beta$-normal form, such that $\forall F_i$ $\in$ $\Theta_F$, $H = G @ F_i$, then $Inverse_{R}(H,G)$ will give an $F$, where $F$ $\in$ $\Theta_F$.
\end{theorem}

\begin{theorem}[$Inverse_L$ complexity]
The $Inverse_L$ Algorithm runs in exponential time in the number of variables in $G$ and polynomial time in the size of the formulas $H$ and $G$.
\end{theorem}

\begin{theorem}[$Inverse_R$ complexity]
The $Inverse_R$ Algorithm runs in exponential time in the number of variables in $G$ and polynomial time in the size of the formulas $H$ and $G$.
\end{theorem}

Due to lack of space, we will only comment on how the soundness and completeness proofs are structured.  The complete proofs are given in the online appendix
of the paper. The soundness proof shows how in each of the four cases of $Inverse_L$, the typed ASP $\lambda$-calculus formula $H$ is obtained by applying $F$ to $G$. The application $F$@$G$ is computed using the expressions from the algorithm for $F$ and $G$, generating the expression for $H$ given in the algorithm. The proof of Theorem 1 is given in the online appendix of the paper, pp. 2--3.  The same reasoning is followed for $Inverse_R$. The complete proof of Theorem 2 is given in the online appendix of the paper, pp. 3--4.

The completeness proof is divided to six cases, which correspond to the six possible valid combinations of orders that $H$, $F$ and $G$ may have, such that the order of the terms will be less than 2. These are shown in Table 4. For each case, it is proven by contradiction that $Inverse_L$ and $Inverse_R$ return a formula $F$ if one such $F$ exists. It is done by assuming that they return a $null$ value and reaching a contradiction at the end of the proof. In the process, each of the four conditions of the algorithms are analyzed, where it is shown that at least one of the conditions of the algorithm has be satisfied for each of the six cases.  The complete proof of Theorem 3 is given in the online appendix of the paper, pp. 4--7 and Theorem 4 is given in pp. 7--11.

Finally, the proof for the complexity results, Theorem 5 and Theorem 6, are given in the online appendix of the paper, pp. 11--12.

\begin{table}[htb]
\footnotesize{
\begin{center}
\begin{tabular}{c c c | r c l}
H & F & G & \multicolumn{3}{c}{ASP type examples for formula F}\\
\hline
0 & 1 & 0 & $e$ & $\rightarrow$ &  $t$\\

1 & 1 & 0 & $a$ & $\rightarrow$ & $(e \rightarrow t)$\\

2 & 2 & 0 & $d$ & $\rightarrow$ & $((g \rightarrow t) \rightarrow t)$\\

0 & 2 & 1 & $(h \rightarrow t)$ & $\rightarrow$ & $t$\\

1 & 2 & 1 & $(l \rightarrow t)$ & $\rightarrow$ & $(e \rightarrow t)$\\

2 & 2 & 1 & $(g \rightarrow t)$ & $\rightarrow$ & $((e \rightarrow t) \rightarrow t)$\\
\end{tabular}
\end{center}
}
\caption{Possible order combinations for F, G and H formulas, with $H=F@G$.}
\label{tab:Forders}
\end{table}

\section{Conclusion}

In this paper we developed the language of typed answer set programming lambda calculus and defined associated notions such as ASP typed term, ASP $\lambda$-calculus formulas, and type orders. We used these notions to formulate soundness and completeness of Inverse $\lambda$-Algorithms  with respect to  typed answer set programming lambda calculus. These algorithms are important in that they allow automatic construction of ASP-lambda representations of new words using information already available about known sentences and words. They have been used in a system that is able to learns to translate combinatorial logic puzzle descriptions to ASP rules \cite{baral2012-kr} that obtain solutions to the puzzles; however that (short) paper does not go into the details of the algorithm, as we do here. 


\section*{Acknowledgement}

We acknowledge support by NSF (grant number 0950440), IARPA and ONR-MURI for this work.

\newpage

\bibliographystyle{acmtrans}
\bibliography{biblio}

\end{document}